\documentclass{article}
\usepackage{nips15submit_e,times}

\usepackage{latexsym}
\usepackage{graphicx}
\usepackage{algorithm}
\usepackage{algorithmic}
\usepackage{amsmath}

\title{Generating Text with Deep Reinforcement Learning}

\author{
Hongyu Guo%\thanks{ Use footnote for providing further information about author (webpage, alternative address)---\emph{not} for acknowledging funding agencies.} 
\\
National Research Council Canada\\
\texttt{hongyu.guo@nrc-cnrc.gc.ca} 
}

\nipsfinalcopy % Uncomment for camera-ready version

\begin{document}

\maketitle
%\vspace{-3.5mm}
\begin{abstract}
%\vspace{-2.5mm}
We introduce a novel schema for sequence to sequence learning with a Deep Q-Network (DQN), which  decodes the output sequence {\em iteratively}. The aim here is to enable the decoder to  first tackle easier portions of the sequences, and then turn to cope with difficult parts.  
Specifically, in each iteration, an encoder-decoder Long Short-Term Memory (LSTM) network is employed to, from the input sequence, automatically create features to represent the internal states of and formulate a  list of potential actions for the DQN. Take rephrasing a natural sentence as an example. This list can contain ranked potential words. Next, the DQN learns to make decision on which action (e.g., word) will be selected from the list to modify the current decoded sequence. 
The newly modified output sequence is subsequently used as the input to the DQN for the next decoding iteration.  
 In each iteration,  we also bias the reinforcement learning's attention to explore  sequence portions which are previously difficult to be decoded. 
 For evaluation, the proposed strategy was trained to  decode ten thousands natural sentences. 
Our experiments indicate that, when compared to a left-to-right greedy beam search LSTM decoder, the proposed method performed competitively well when decoding sentences from the training set, but  significantly outperformed the baseline when  decoding unseen sentences, in terms of BLEU score obtained.
\end{abstract}
\vspace{-3.5mm}
\section{Introduction}
\vspace{-2.5mm}
Many  real-world problems can be effectively  formulated as sequence to sequence learning.   Important applications include speech recognition, machine translation, text rephrasing,  question
answering. For example, the last three can  be expressed as mapping a sentence of words to another  sequence of words. A major challenge of modeling these tasks is the variable length of sequences  which is often not known a-priori. To address that, an encoder-decoder Long Short-Term Memory (LSTM) architecture has been recently shown to be very effective~\cite{Cho:2014,Sutskever:2014}. 
The idea is to use one LSTM to encode the input sequence, resulting in a fixed dimensional
vector representation. Subsequently, another LSTM is deployed to decode (generate) the output sequence, using  the newly created vector  as the LSTM's initial state. The decoding process is  essentially a recurrent neural network language model~\cite{mikolov2010recurrent,sundermeyer2012lstm}.  

Decoding schema based on recurrent language models naturally fits a left-to-right decoding procedure, which aims to  obtain an output sequence with the maximal  probability or to select the top list of sequence candidates for further post-processing. In this paper, we propose an alternative strategy for training an end-to-end decoder. Specifically, we employ a Deep Q-Network (DQN) to embrace an iterative decoding strategy. 
In detail,  
the input sequence is first encoded using an encoder-decoder LSTM network. 
 This process  automatically generates both informative features to represent the internal states of and a  list of potential actions for a DQN.  
Next, the DQN is employed to  {\em iteratively} decode the output sequence.  Consider rephrasing a natural sentence.  This list of potential actions can contain the ranked word candidates. In this scenario, the DQN learns to make decision on which word will be selected from the list to modify the current decoded sequence. 
The newly edited  output sequence is subsequently used as the input to the DQN for the next decoding iteration. 
 Inspired by the recent success of attention mechanisms~\cite{BahdanauCB14,GregorDGW15,MnihHGK14,SukhbaatarSWF15}, we here also bias the reinforcement learning's attention, in each iteration, to explore  sequence portions which are previously difficult to be decoded.  
The decoded sequence of the last iteration is used as the final output of the model. 
In this way, unlike the left-to-right decoding schema, the DQN is able to learn to first focus on the easier parts of the sequence, and  the resulted new information  is then use to help solve the difficult portions of the sequence.  
For example, a sentence from our testing data set was decoded by the encoder-decoder LSTMs as ``Click here to read more {\em than} the New York Times .'', which was successfully corrected by the DQN as ``Click here to read more {\em from} the New York Times .'' in the second iteration. 

For evaluation, the proposed strategy was trained to encode and then decode  ten thousands natural sentences.  
Our experimental studies indicate that the proposed method performed competitively well for decoding sentences from the training set, when compared to a left-to-right greedy beam search decoder with LSTMs, but  significantly outperformed  the baseline when  decoding unseen sentences, in terms of BLEU~\cite{Papineni:2002} score obtained.  

 Under the context of reinforcement learning, decoding sequential text  will  need to overcome the challenge arise from the  very large number of potential states and actions.   This is mainly due to the  flexible  word  ordering of a sentence and the existence of a  large  number  of   words and synonyms in modern dictionaries. 
 To our best knowledge, our work is the first to decode text using DQN. In particular, we employ LSTMs to not only  generalize informative features from text to represent the states of DQN, but also create a list of potential actions (e.g., word candidates) from the text for the DQN. Intuitively, the application of the  DQN here also has the effect of generating \textit{synthetic} sequential text for the training of the networks, because of the DQN's exploration strategy in training. 

\vspace{-2.5mm}
\section{Background}
\vspace{-2.5mm}
\textbf{Reinforcement Learning and Deep Q-Network} 
Reinforcement Learning (RL) is a commonly used framework for learning control policies by a computer algorithm, the so-called agent, through interacting with its environment $\Xi$~\cite{Amato:2010,Silver:2007}. 
 Given a set of internal states  $S={s_{1}, \dots,s_{I}}$ and a set of predefined actions $A={a_{1}, \dots, a_{k}}$, 
 the agent takes action $a$ at state $s$, by following certain policies or rules,  will result in a new state $s^{,}$ and receive a reward $r$ from $\Xi$. The aim of the agent is to maximize some cumulative reward through a sequence of actions. Each such action forms a transition tuple $(s, a, r, s^{,})$ of a Markov Decision Process (MDP). 
Practically, the environment is unknown or partially observed, and a sequence of state transition tuples can be used to formulated the environment. 
   
Q-Learning~\cite{Watkins92qlearning} is a popular form of RL. This model-free technique is used to learn an optimal action-value function Q(s, a), a measure of the action's expected long-term reward,  for the agent. 
 Typically, Q-value function relies on all possible state-action pairs, which are often impractically to be obtained. A work around for this challenge is to  approximate Q(s, a) using a parameterized function $Q(s, a; \theta)$. The parameter $\theta$ is often learned by features generalized over the states and actions of the environment~\cite{Branavan2011,Sutton1998}. 
  Promisingly, benefiting from the recent advance in deep learning techniques, which have shown be able to effectively  generate informative features for a wide ranges of difficult problems, Mnih et al.~\cite{mnihdqn2015} introduced the Deep Q-Network (DQN). The DQN approximates the Q-value function with a non-linear
 deep convolutional network, which also automatically creates  useful features to represent the internal states of the RL.  

In DQN, the agent interacts with environment $\Xi$ in discrete iteration $i$, taking aim to maximize its long term reward. 
Starting from a random Q-function, the agent continuously updates its Q-values by taking actions and obtaining rewards, through consulting a current Q-value function. 
   The iterative updates are derived from the Bellman equation, where  the expectation $E$ is often computed over all transition tuples that involved the agent taking action $a$ in state $s$~\cite{Sutton1998}: 
\begin{align}
Q_{i+1}(s, a) = E[r + \lambda\underset{a^{,}}{\operatorname{max}} Q_{i}(s^{,}, a^{,}|s,a)
\label{bell}
\end{align}
Where $\lambda$ is a discounted factor for future rewards. 

DQN requires informative representation of internal states. For playing video games, one can infer state representations directly from raw pixels of screens using a convolutional network~\cite{mnihdqn2015}. However, text sentences, for instance, not only contain  sequential nature of text, but also have variable length.  The LSTM's ability to  learn on data with long range temporal dependencies and varying lengths makes it a natural choice to replace the convolutional network in the DQN for our application here. Next, we will briefly describe the LSTM network.

\textbf{Long Short-Term Memory Recurrent Neural Networks} 
Through deploying a recurrent hidden vector, Recurrent Neural Networks (RNNs)~\footnote{We here describe the commonly used Elman-type RNNs~\cite{Elman90findingstructure}; other variants such as Jordan-type RNNs~\cite{Jordan1997471} are also available in the community.} can compute compositional vector representations for  sequences of arbitrary length. The network  learns complex temporal dynamics by mapping  a length $T$ input sequence $<x_{1}, x_{2}, \dots, x_{T}>$ to a sequence of hidden states $<h_{1}, h_{2}, \dots, h_{T}>$ ($h_{t} \in R^{N}$).  
The networks compute the hidden state vector via the recursive application of a transition function:  
\begin{align}
h_{t} &= \Gamma (W_{xh}x_{t} +W_{hh}h_{t-1} + b_{h}) \label{rnn}
\end{align}
where $\Gamma$ is an element-wise non-linearity sigmoid function; the $W$ terms denote weight matrices (e.g. $W_{xh}$ is the input-hidden weight matrix);  $b_{h}$ is hidden bias vector.

A popular variant of RNNs, namely LSTMs are  designed to overcome the vanishing gradient issue in RNNs, thus better modeling long term dependencies in a sequence.  
In addition to a hidden unit $h_{t}$, LSTM includes input gate, forget gate, output gate and memory cell unit vectors,  for the following purposes. 
 The memory cell unit $c_{t}$, with a self-connection, is capable of considering two pieces of information. The first one is the previous memory cell unit $c_{t-1}$, which is modulated by the forget gate. Here, the forget gate  embraces the hidden states to adaptively reset its cell unit through the self-connection.  
 The second piece of information is a function of the current input and previous hidden state, modulated by the input gate. 
Intuitively, the LSTM can learn to selectively forget its previous memory or consider its current input. Similarly, the output gate  learns how much of the memory cell to transfer to the hidden
state. These additional cells enable the LSTM to preserve state over long periods of time~\cite{Cho:2014,Graves:2013,Sutskever:2014,Vinyals:2014}.

\begin{figure}[h]
  \centering
  \includegraphics[width=3.3in]{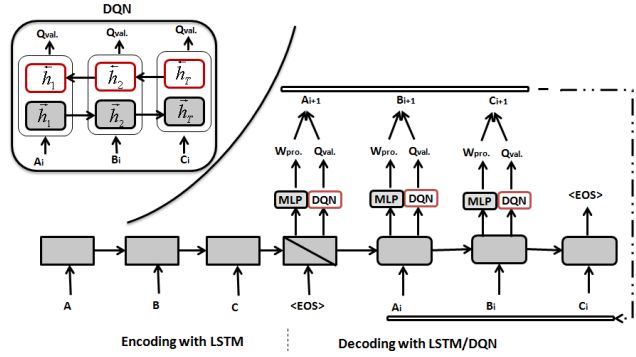}
  \caption{Iteratively decoding with DQN and LSTM; 
  the encoder-decoder LSTM network is depicted as gray-filled rectangles on the bottom; 
  the top-left is the graphical illustration of the DQN with bidirectional LSTMs; the dash arrow line on the right indicates the iteration loop.}
  \label{fig:dqn}
\end{figure}

\vspace{-3.5mm}

\section{Generating Sequence with Deep Q-Network}
\vspace{-2.5mm}
We employ an encoder-decoder LSTM network, as presented in~\cite{Sutskever:2014},  to automatically generate informative   features for a DQN, so that the DQN can   learn a Q-value function to approximate its long term rewards. The learning algorithm is depicted in Figure~\ref{fig:dqn} and Algorithm~\ref{alg1}.

\vspace{-2.5mm}
\subsection{Generating  State Representations with LSTMs}
\vspace{-2.5mm}
 The encoder-decoder LSTM network is depicted as gray-filled rectangles in Figure~\ref{fig:dqn}.  
  For descriptive purpose, we named this State Generation Function (denoted as StateGF) under the context of DQN. 
In detail, given a natural sentence with $N$ tokens, $<x_{1}, x_{2}, \dots, x_{N}>$ (denoted as EnSen). We first encode the sequence using one LSTM (denoted as EnLSTM), reading into the tokens (e.g., words) one timestep at a time (e.g., $<A,B,C>$ in Figure~\ref{fig:dqn}). When reaching the end of the sentence ($<$EOS$>$ in Figure~\ref{fig:dqn}), this encode process results in a  fixed dimensional vector representation for the whole sentence, namely the hidden layer vector $h_{N}^{en}$.  
Next, the resulted $h_{N}^{en}$ is used as the initial state of another LSTM (denoted as DeLSTM) for decoding to generate the target sequence $<y_{1}, y_{2}, \dots, y_{T}>$.  
In this process, the hidden vectors of the DeLSTM are also conditioned on its input (i.e., $<A_{i},B_{i},C_{i}>$ in Figure~\ref{fig:dqn}; for a typical language model, this will be $<y_{1}, y_{2}, \dots, y_{T}>$). 
 Consequently, the DeLSTM creates a sequence of hidden states $<h_{1}^{de}, h_{2}^{de}, \dots, h_{T}^{de}>$ ($h_{t}^{de} \in R^{N}$) for each time step. 
Next, each of these hidden vectors is fed into a {\em Softmax} function to 
produce a distribution over the $C$ possible classes (e.g., words in a vocabulary or dictionary), thus creating a list of word probabilities at  each time step $t$, i.e., $<W_{pro1}^{t}, W_{pro2}^{t}, \dots, W_{proV}^{t}>$ (V is the size of the dictionary): 
\begin{align}
\small
P(W_{pro}^{t} = c|EnSen,\vartheta) =\frac{exp (w^{T}_{c}h_{t}^{de})}{\sum^{C}_{c=1} exp (w^{T}_{c}h_{t}^{de})}
\end{align}
 where $w_{c}$ is the weight matrix from the  hidden layer to the output layer. 
 These probabilities can be further processed by a $Argmax$ function, resulting in a sequence of output words, namely a sentence $<y^{i}_{1}, y^{i}_{2}, \dots, y^{i}_{T}>$ (denoted as DeSen$_{i}$; $i$ indicates the $i$-th iteration of the DQN, which will discussed in detail later).

The parameter $\vartheta$ for the decoder-encoder LSTMs, namely the StateGF function, is tuned  to 
maximize the log probability of a correct decoding sentence $Y$ given the source sentence $X$, using the following training objective:
\begin{align}
1/|S| \sum^{}_{(X,Y) \in S} log p(Y|X)
\end{align}
where $S$ is the training set. After training, decoding output sequence can be achieved  by finding the most likely output sequence according to the DeLSTM:
\begin{align}
 \hat{Y} =  \underset{Y}{\operatorname{argmax}} p(Y|X)
\end{align}
%\vspace{-2.5mm}
A straight forward and effective method for this decoding search, as suggested by~\cite{Sutskever:2014}, is to deploy a simple left-to-right beam search. That is, the decoder  maintains a small number of incomplete sentences. At each timestep, the decoder extends each partial sentence in the beam with every possible word in the vocabulary.  
As suggested by~\cite{Sutskever:2014}, a  beam size of 1 works well.  

In this way, feeding the  state generate function with EnSen will result in a decoded sentence DeSen. 
The DQN decoding, which will be discussed next, employs an iteration strategy, so we denote this sentence sequence   pair as EnSen$_{i}$  and DeSen$_{i}$; here $i$ indicates the $i$-th iteration of the DQN. 

\begin{algorithm} [h]
\caption{Generating Text with Deep Q-Network}
%\small
%\footnotesize
\scriptsize
\label{alg1}
\algsetup{indent=2em}
\begin{algorithmic} [1]
\STATE Initialize replay memory $D$; initialize EnLSTM, DeLSTM, and DQN with random weights
%\STATE
\STATE \textbf{Pretraining  Encoder-Decoder LSTMs}
\FOR {epoch = 1,M}
\STATE randomize given training set with sequence pairs $<X,Y>$.
\FOR {each sequence pair $EnSen^{k} \in X$ and $TaSen^{k} \in Y$}
\STATE Encode $EnSen^{k}$ with EnLSTM, and then predict the next token (e.g., word) in $TaSen^{k}$ with DeLSTM.
\ENDFOR
\ENDFOR
%\STATE \textbf{Pretraining Ends}
%\STATE 
\STATE \textbf{Training Q-value function}
\FOR {epoch = 1,U}
\FOR {each sequence pair $EnSen^{k} \in X$ and $TaSen^{k} \in Y$ (with length $l$)}
\STATE feed $EnSen^{k}$ into  pretrained encoder-decoder LSTMs; obtain the decoded sequence $DeSen^{k}_{0}$
\FOR {iteration i = 1, $2l$}
\IF {random() $ < \epsilon$} 
\STATE select a random action $a_{t}$ (e.g., word $w$) at time step $t$ of $DeSen^{k}_{i}$ (selection biases to  incorrect decoded  tokens)
\ELSE 
\STATE compute $Q(s_{i},a)$ for all actions using DQN; select $a_{t}=argmax Q(S_{i},a)$, resulting in a new token $w$ for the $t$-th token in $DeSen^{k}_{i}$
\ENDIF
\STATE replace $DeSen_{i}^{k}$ with $w$, resulting $DeSen_{i+1}^{k}$
\STATE compute the similarity of $DeSen_{i+1}^{k}$ and $TaSen^{k}$, resulting reward score $r_{i}$%SmoothedBLEU score $\vartheta$, then thus 
\STATE store transition tuple [$s_{i},a_{i},r_{i},s_{i+1}$] in replay memory $D$; $s_{i}=[EnSen^{k}_{i},DeSen^{k}_{i}]$. 
\STATE random sample  of transition [$s_{i},a_{t},r_{i},s_{i+1}$] in $D$
\IF {$r_{i} > \sigma$ (preset BLEU score threshold)} 
\STATE $q_{i}=r_{i}$; current sequence decoding successfully complete. 
\ELSE 
\STATE $q_{i} = r_{i} +  \lambda max_{a^{,}} Q(s^{,}, a^{,}; \theta_{i-1})$
\ENDIF
\STATE  perform gradient descent step on only the DQN network
 $(q_{i} - Q(s, a; \theta_{i}))^{2}$
\ENDFOR
\ENDFOR
\ENDFOR
%\STATE \textbf{Training Q-value function ends}
\end{algorithmic}
\end{algorithm}

\vspace{-2.5mm}
\subsection{Iteratively Decoding Sequence with Deep Q-Network }
\vspace{-2.5mm}
At each decoding iteration $i$, the DQN considers the sentence pair, namely the EnSen$_{i}$  and DeSen$_{i}$ (i.e., $<A,B,C>$ and $<A_{i}, B_{i}, C_{i}>$, respectively, in Figure~\ref{fig:dqn}) as its internal state. Also, the ranked words list~\footnote{This list has the same size as that of the vocabulary; but one can   use only a few of the top ranked words.} for each time step $t$ of the DeLSTM is treated as the potential actions by the DQN. 
From these lists, 
the DQN learns to predict what actions should be taken in order to accumulate larger long time reward. 

In detail, each hidden vector $h_{t}^{de}$ in the DeLSTM is  fed into a neural network (depicted as DQN in figure~\ref{fig:dqn} and graphically illustrated on the top-left subfigure; will be further discussed in Section~\ref{bidir}). These neural networks learn to approximate the Q-value function given the DQN's current state, which contains the EnSen$_{i}$  and DeSen$_{i}$ as well as the word probability list at each time step $t$ of the DeLSTM.  The DQN will take the action with the max Q-value in the outputs. Consider, the DQN takes an action, namely selects the $t$-th time step word $\widehat{y^{i}_{t}}$ in iteration $i$. Then the current state of the DQN will be modified accordingly. That is, the DeSen$_{i}$ will be modified by replacing the word at time step $t$, namely replacing $y_{t}^{i}$ with $\widehat{y^{i}_{t}}$. This process results in a new decoded sentence, namely DeSen$_{i+1}$ (depicted as $<A_{i+1}, B_{i+1}, C_{i+1}>$ in Figure~\ref{fig:dqn}). Next, the similarity of the target sentence  $<y_{1}, y_{2}, \dots, y_{T}>$ and the current decoded sentence DeSen$_{i+1}$ is evaluated by a BLEU metric~\cite{Papineni:2002}, which then assigns a reward $r_{i}$ to the action of selecting $\widehat{y^{i}_{t}}$.   
Thus, a transition tuple for the DQN contains  [(EnSen$_{i}$, DeSen$_{i}$), $\widehat{y^{i}_{t}}$, $r_{i}$, ([EnSen$_{i}$, DeSen$_{i+1}$]). In the next iteration of the DQN,  the newly generated sentence DeSen$_{i+1}$  is then fed into the DQN to generate the next decoded sentence DeSen$_{i+2}$.

The training of the DQN is to find the optimal weight matrix  $\theta$ in the neural networks. That is,  the Q-network is trained by minimizing a sequence of loss functions $L_{i}(\theta_{i})$ at each iteration $i$:
\begin{align}
L_{i}(\theta_{i}) = E_{s,a}[(q_{i} - Q(s,a; \theta_{i}))^{2}]
\end{align}
where $q_{i} = E_{s,a} [r_{i} +  \lambda max_{a^{,}} Q(s^{,}, a^{,}; \theta_{i-1}) | s, a]$
is the target Q-value, or reward, with parameters $\theta_{i-1}$ fixed from the previous iteration.
In other words, the DQN is trained to predict its expected future reward. 
The updates on the parameters $L_{i}(\theta_{i})$ is performed with the following gradient: %  $L_{i}(\theta_{i})$
\begin{align}
\nabla_{\theta_{i}} L_{i}(\theta_{i}) = E_{s,a}[2(q_{i} - Q(s, a; \theta_{i})) \nabla_{\theta_{i}} Q(s, a; \theta_{i})]
\end{align}
After learning the Q-value function, the agent chooses the action with the highest $Q(s, a)$ in order to maximize its expected future rewards when decoding sequences. 
Quite often, a  trade-off between exploration and exploitation strategy is employed for the agent. That is, through  following an $\epsilon$-greedy policy, the agent can perform a random action with probability $\epsilon$~\cite{Sutton1998}. 
Inspired by the recent success of attention mechanisms~\cite{BahdanauCB14,GregorDGW15,MnihHGK14,SukhbaatarSWF15},  we here bias the reinforcement learning's attention to explore the sequence portions which are difficult to be decoded. That is, the random actions have more chance to be picked for tokens which were  decoded incorrectly from the previous iterations.

\vspace{-2mm}
\subsection{Bidirectional LSTMs for DQN}
\label{bidir}
\vspace{-2mm}
 During decoding, we would like the DQN to have information about the entire input sequence, i.e., 
$<A_{i}, B_{i}, C_{i}>$ in Figure~\ref{fig:dqn}. To attain this goal, we deploy a bidirectional LSTMs~\cite{Graves:2013}. 
Specifically, for a specific time step $t$ of a given sequence, a Bidirectional LSTM~\cite{Graves:2013} enables the hidden states   to  summarize time step $t$'s past and future in the sequence. The network deploys two separate hidden layers to precess the data in both directions: one from left to right (forward), and another right to left (backward). At each time step, the hidden state of the Bidirectional LSTM is the concatenation of the forward and backward hidden states, and then fed forwards to the same output layer. That is, Equation~\ref{rnn} for the DQN is implemented as follows (illustrated in the top-left subfigure in Figure~\ref{fig:dqn}). 
\begin{align}
\footnotesize
\overrightarrow{h_{t}} &= \Gamma (W_{x\overrightarrow{h}}x_{t} +W_{\overrightarrow{h}\overrightarrow{h}}\overrightarrow{h_{t-1}} + b_{\overrightarrow{h}}) \\
\overleftarrow{h_{t}} &= \Gamma (W_{x\overleftarrow{h}}x_{t} +W_{\overleftarrow{h}\overleftarrow{h}}\overleftarrow{h_{t-1}} + b_{\overleftarrow{h}}) \\
h_{t} &= [\overrightarrow{h_{t}^{T}} ; \overleftarrow{h_{t}^{T}}]^{T}
\end{align}
In this scenario, $h_{t}^{de}$ is equal to $\overrightarrow{h_{t}}$, namely the forward hidden vectors. The additional information about the input sequence $<A_{i}, B_{i}, C_{i}>$ is further summarized by the backward hidden vectors $\overleftarrow{h_{t}}$.  

\vspace{-2mm}
\subsection{BLEU Score for DQN Reward}
\vspace{-2.5mm}
Reward is calculated based on the closeness between the  target sentence  $<y_{1}, y_{2}, \dots, y_{T}>$ and the decoded output sentence (i.e., $DeSen$) after the DQN takes an action. We compute the similarity of this sentence pair  using 
the  popular score metric in statistical translation. Specifically, we obtain a BLEU~\cite{Papineni:2002} score between these two sentences. We measure the score difference between the current iteration and the previous iteration. If the difference is positive, then a reward of  +1 is assigned; if negative then -1 as reward; otherwise, it is zero. 
Note that, 
 since we here conduct a sentence level comparison, we adopt the smoothed version of BLEU~\cite{Lin:2004:AEM:1218955.1219032}. Unlike the BLEU, the smoothedBLEU avoids giving zero score even when there are not any 4-gram matches in the sentence pair.

\vspace{-1.95mm}
\subsection{Empirical Observations on Model Design}
\vspace{-1.95mm}
%\subsection{Discussions on Model Design}
\textbf{Separating State Generation Function from DQN} 
Our experiments suggest that separating the state generation function from the DQN networks is beneficial. The aim here is to  have a deterministic network for generating states from a  sequence pair. That is, for any given input pair to the encoder-decoder LSTMs network, namely the state generation function StateGF, we will always have the same decoded output sequence. Our empirical studies indicate that this is a very important for successfully training the DQN for decoding text. Our intuitive explanation is as follows. 

  Using DQN to approximate the Q-value function, intuitively, equals to train a network against moving targets because here the network's targets depend on the network itself. 
  Suppose, for a given input feed, the StateGF would generate a different output sequence each time for the DQN. In this scenario,  the DQN network has to also deal with a moving state function involving text with \textit{very high dimensionality}. Intuitively, here the DQN agent is living in a changing environment  $\Xi$.  
 As a result,  it may be very difficult for the DQN to learn to predict the Q-value, since, now, all the states and rewards are unstable, and change even for the same input feed. 
 
\textbf{Pre-training the State Generation Function}
Two empirical techniques are employed to ensure that we have a deterministic network for generating states for DQN.
Firstly, we deploy a pre-training technique. Specifically, we pre-train the state generation function StateGF with  the input sequence $X$  as the EnLSTM's input and target sequence $Y$ as the DeLSTM's input. After the training converges, the networks' weights will be fixed  when the training of the DQN network starts. 
Secondly, during training the DQN, the input sequence is fed into the EnLSTM, but the  decoded sequence from the previous iteration is used by  the DeLSTM as input (indicated as dot line in Figure~\ref{fig:dqn}). In this stage, only the red portions of Figure~\ref{fig:dqn} are updated. That is, the reward errors from the DQN networks are not backpropagated to the state generation functions.

\textbf{Updating with Replay Memory Sampling}
Our studies also indicate that, performing updates to the Q-value function using transitions from the current training sentence causes the network to strongly overfit the current input sentence. As a result, when a new sentence is fed in for training, it may always predict the previous sentence used for training. To avoid this correlation issue, a replay memory strategy is applied when updating the DQN. That is, the DQN is updated by transition tuples  which may be  different from the current input sequence. 

To this end, for each action the DQN takes, we save its transition tuple in the replay memory pool, including the EnSen$_{i}$, DeSen$_{i}$, DeSen$_{i+1}$,  $r_{i}$, and  $a_{i}$. When updating the DQN, we then randomly sample a  transition tuple from the replay memory pool. 
More sophisticated replay memory update could be applied here; we would like to leave it for future work. For example, one can use the  priority sampling of replay  technique~\cite{Moore93prioritizedsweeping}. That is,  transitions with large rewards have more chance to be chose.  In our case, we can bias the selection to transitions with  high BLEU scores.

\textbf{Importance of Supervised Softmax Signal}
We also conduct experiments without the supervised $Softmax$ error for the network. That is, the whole network, including  the LSTMs and DQN, only receive the error signals from the Q-value predictions. We  observed that, without the supervised signal the DQN was very difficult to learn. The intuition is as follows.  Firstly, as discussed before, for decoding text, which typically involves a very large number of potential states and actions, it is  very challenge for the DQN to learn the optimal policy from both a moving state generation function and a moving Q-value target function. Secondly, the potential actions for the DQN, namely the word probability list for each output of the DeLSTM is changing and unreliable, which will further complicate the learning of the DQN. 

\textbf{Simultaneously Updating with Both Softmax and Q-value Error}
If during training the DQN, we not only update the DQN as discussed previously, but also update the state generation functions, i.e., the encoder-decoder LSTMs. We found that the network could be easily bias to the state generation functions since the $Softmax$ error signal is very strong and more reliable (compared to the moving target Q-value function), thus the DQN may not be  sufficiently tuned. Of course, we could bias towards the learning of DQN, but this would introduce one more tricky parameter for tuning. In addition, doing so, we have an indeterministic state generation function again.   

\vspace{-2.4mm}
\section{Experiments}
\vspace{-3.4mm}
\subsection{Task and Dataset}
\vspace{-2.5mm}
Our experimental task here is to train a network to regenerate natural sentences. That is, given a sentence as input, the  network  first compresses it into a fixed vector, and then this vector is used to decode the input sentence. In other words, the $X$ and $Y$ in Algorithm~\ref{alg1} are the same. 
In our experiment, we randomly select 12000 sentences, with max length of 30, from 
 the Billion Word Corpus~\cite{ChelbaMSGBKR14}. We  train our model with 10000 sentences, and then select the best model with the  validation data which consist of 1000 sentences. We then test our model with 1000 seen sentences and 1000 unseen sentences. The seen test set is randomly sampled from the training set. 
 
 \vspace{-2.5mm}
\subsection{Training and Testing Detail}
\vspace{-2.5mm}
%\subsection{Network Initialization}
For computational reason, we used an one-layer LSTM for the encoder-decoder LSTMs as well as the backward LSTM in the DQN, both with 100 memory cells and 100 dimensional word embeddings. 
  We used a  Softmax over 10000 words (which is the size of the vocabulary we used in the experiments) at each output (i.e., time step $t$) of the DeLSTM.
We initialized all of the LSTM's parameters with the uniform distribution between  -0.15 and +0.15, including the word  vectors. 
We used Adaptive Stochastic Gradient Descent (AdaSGD)~\cite{Duchi:2011} without momentum, with a starting learning rate of 0.05. 
 Although LSTMs tend to not suffer from the vanishing gradient problem, they can have
exploding gradients. Thus we employ the gradient norm clip technique~\cite{PascanuICML2013} with a threshold of 15. We used both L2 regularization (with a weight decay value of 0.00016) and dropout (with a rate of 0.2) to avoid overfitting the networks. 

%\subsection{Training and Testing Detail}
We first pretrain the encoder-decoder LSTMs with both the target sentence as input. After the training converges, we then start to train the DQN. 
When training the DQN, we turn off the drop out in the encoder-decoder LSTMs, so that we have a deterministic network to generate states and the lists of word probabilities for  the DQN. 
In addition, we scale down the epsilon $\epsilon$ to 0.1 after 2000000 iterations. In other words, most of actions at the beginning of the DQN training were  random, and then became more greedy towards the end of the training. 
For each sentence with length of $l$, we allow  DQN to edit the sentence with $2l$ iterations, namely taking $2l$ actions for the decoding. The sentence decoded in each iteration will be saved in a replay memory  with a capacity of 500000. The discount factor $\lambda$ was set to 0.95. 
Also, the BLEU score threshold $\sigma$ for indicating  decoding success was set to 0.92. 
 For the initial states of the bi-directional LSTMs in the DQN, we used the fixed vector generated by the LSTM encoder. 

In testing phase, we also run the DQN for each sentence with $2l$ steps. 
Also, in our experiment, we used only the word with the max probability on each of the $T$ lists as the potential actions for the DQN. Since the maximal length of a sentence in our experiment is 30, the DQN has at most 31 output nodes. Namely, the DQN can choose one of the 30 top words, each corresponding to a time step at the DeLSTM, as its action, or take the 31$^{st}$ action which indicates not modification is needed. 

%\subsection{Baseline and Evaluation}
We compared our strategy with an encoder-decoder LSTM network used in~\cite{Sutskever:2014} for machine translation. This baseline decoder 
 searches for the most likely output sequence using a simple left-to-right beam search technique.  
As suggested by~\cite{Sutskever:2014}, a  beam size of 1 worked well. We adopt this approach as our decoding baseline. 
All our experiments were run on a NVIDIA GTX TitanX GPU with 12GB memory. 
We report the average SmoothedBLEU score for all the testing sentences. 

\vspace{-1.9mm}
\subsection{Experimental Results}
\begin{figure}[h]
  \centering
  \includegraphics[width=4.25in]{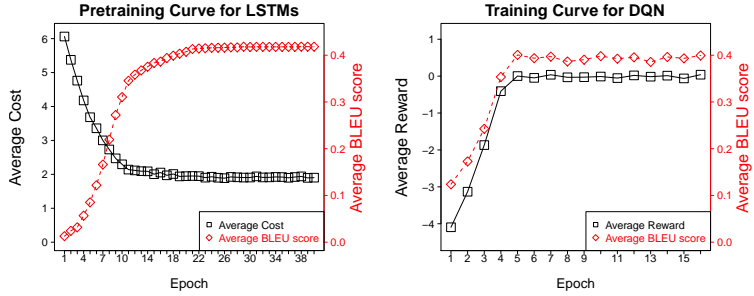}
  \caption{The evolution of   cost for training the StateGF  and   reward for training the DQN.}
  \label{fig:train}
\end{figure}

\vspace{-1.9mm}
The evolutions of the training for the state generation function StateGF and DQN are depicted in Figure~\ref{fig:train}, and the main testing results are presented in Table~\ref{tab:test}.

%\vspace{-1.9mm}
%\vspace{-1.9mm}

From the  training curves for both the encoder-decoder LSTMs and DQN as depicted in Figure~\ref{fig:train}, we can see that both of the trainings converged very well. For the LSTMs training, the average cost was steadily decreasing, and the average BLEU score was gradually increasing. Both the two curves then stabilized after 20 iterations. Also, the right subfigure in Figure~\ref{fig:train} indicates that, the reward obtained by the DQN was negative at the beginning of the training and then gradually moved to the positive reword zone. The negative rewards are expected here because most of the DQN's actions were random at the beginning of the training. Gradually, the DQN knew how to decode the sentences to receive positive rewards. As shown in the right figure, the training of the DQN converged in about 6 epochs. These results indicate that both the state generation functions and the DQN were easy to be trained.

\begin{table*}[h]
  \centering
\begin{tabular}{l|c|c}\hline
Testing systems   & LSTM decoder  & DQN        \\
\hline \hline
Average SmoothedBLEU on sentences IN the training set&0.425&0.494 \\
Average SmoothedBLEU on sentences NOT in the training set                          & 0.107&  0.228   \\\hline
\end{tabular}
  \caption{Experimental results for decoding the seen and unseen sentences in testing.}
  \label{tab:test}
\end{table*}

In Table~\ref{tab:test}, we show the testing results, in terms of average SmoothedBLEU obtained, for both the seen 1000 and unseen 1000 sentences. We can observe that, although the results achieved by the DQN on the seen data were only slightly better than that of  the baseline LSTMs network, for the unseen data the DQN meaningfully outperformed the baseline. Our further analysis suggests the follows. 
With the seen data, the DQN decoder tended to agree with the LSTM decoder. That is, most of the time, its decision was ``no modification''. 
As for the unseen data, because the DQN's exploration strategy allows it to learn from many more noisy data than the LSTMs networks did, so the DQN decoder was able to tolerate better to  noise and generalize well to unseen data.  Intuitively, the application of the  DQN here also has the effect of generating synthetic sequential text for the training of the DQN decoder, due to its exploration component.

\vspace{-1.9mm}
\begin{figure}[h]
  \centering
  \includegraphics[width=1.99201in]{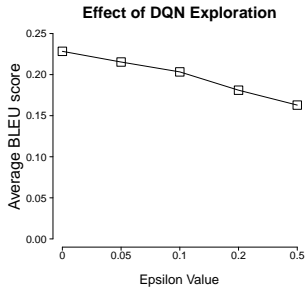}
  \caption{Impact, in terms of BLEU score obtained, of the DQN's exploration in the testing phase.}
  \label{fig:explore}
\end{figure}
\vspace{-1.9mm}

We also conducted experiments to observe the behaviors of the DQN for exploration; here we only considered the unseen testing data set. That is, we enabled the DQN to follow an $\epsilon$-greedy policy with $\epsilon=0, 0.05, 0.1, 0.2, 0.5$, respectively. In other words,  we allowed the agent to choose the best actions according to its Q-value function 100\%, 95\%, 90\%, 80\%, and 50\% of the time. The experimental results, in terms of BLEU score obtained, are presented in Figure~\ref{fig:explore}. From Figure~\ref{fig:explore}, we can conclude that the exploration strategy in testing time did not help the DQN. The results here  indicate that allowing the DQN to explore in testing time decreased its predictive performance, in terms of BLEU score obtained

\section{Related Work}
\vspace{-1.9mm}
Recently,  the Deep Q-Network (DQN) has been shown to be able to successfully play 
  Atari games~\cite{DBLP:journals/corr/HausknechtS15,mnihdqn2015,DBLPjournalsOhGLLS15}. Trained with a variant of Q-learning~\cite{Watkins92qlearning}, the DQN learns control strategies using deep neural networks. The main idea is to use deep learning to automatically generate informative features to represent the internal states of the environment where the software agent lives, and subsequently approximate a non-linear control police function for the learning agent to take actions. 
In addition to playing video games, employing reinforcement learning to learn control policies from text has also be investigated. Applications include interpreting user manuals~\cite{Branavan:2010:RLL:1858681.1858810}, navigating  directions~\cite{artzi2013weakly,Kollar:2010:TUN:1734454.1734553,matuszek2013learning,Vogel:2010:LFN:1858681.1858764} and playing  text-based games~\cite{DBLP:journals/corr/BranavanSB14,Eisenstein:2009:RLC:1699571.1699637,DBLP:journals/corr/NarasimhanKB15}.  Also, DQN has recently been employed to learn memory access patterns and  rearrange a set of given words~\cite{DBLP:journals/corr/ZarembaS15}. 

Unlike the above works, our research here aims to decode natural text with DQN. In addition, we employ an encoder-decoder LSTM network to not only  generalize informative features from text to represent the states of DQN, but also create a list of potential actions from the text for the DQN.

\section{Conclusion and Future Work}
\vspace{-1.9mm}
We deploy a Deep Q-Network (DQN)  to embrace an iterative decoding strategy for sequence to sequence learning. To this end, an encoder-decoder LSTM network is employed to  automatically approximate  internal states and formulate  potential actions for the DQN.  
 In addition, we incorporate an attention mechanism into the reinforcement learning's exploration strategy.  
Such exploration, intuitively, enables the decoding network to learn from many  synthetic sequential text generated during the decoding stage. 
We evaluate the proposed method with a sentence regeneration task. Our experiments demonstrate our approach's  promising performance especially when decoding unseen sentences,  in terms of BLEU score obtained.  
This paper also presents several empirical observations, in terms of model design, in order for successfully decoding sequential text with DQN. 

In the future,  
 allowing
 the DQN to pick from the top $n$ words from the list at each time step $t$ of the DeLSTM would be further studied. Furthermore, we would like to experiment with sophisticated priority sampling techniques for the DQN training. 
 In particular, we are interested in applying this approach to statistical machine translation. 

%\begin{spacing}{0.86}
\small
\bibliographystyle{abbrv}
%\bibliography{reference}

%\end{spacing}

\end{document}